\documentclass{article}
\usepackage{ijcai18}

\usepackage{times}
\usepackage{soul}
\usepackage{url}
\usepackage[hidelinks]{hyperref}
\usepackage[utf8]{inputenc}
\usepackage[small]{caption}

\usepackage{graphicx}
\usepackage{amsmath}
\usepackage{amssymb}
\usepackage{microtype}
\usepackage[ruled]{algorithm2e}
\renewcommand{\footnotesize}{\fontsize{10pt}{12pt}\selectfont}

\makeatletter
\DeclareRobustCommand\onedot{\futurelet\@let@token\@onedot}
\def\@onedot{\ifx\@let@token.\else.\null\fi\xspace}

\def\ie{\emph{i.e}\onedot}

\def\etal{\emph{et al}\onedot}
\makeatother

\title{CR-GAN: Learning Complete Representations for Multi-view Generation}

\author{
Yu Tian$^1$, 
Xi Peng$^1$, 
Long Zhao$^1$, 
Shaoting Zhang$^2$ {\normalfont and} 
Dimitris N. Metaxas$^1$
\\ 
$^1$ Rutgers University\\
$^2$ University of North Carolina at Charlotte\\
\{yt219, px13, lz311, dnm\}@cs.rutgers.edu, szhang16@uncc.edu \\
}

\begin{document}

\maketitle
\begin{abstract}
Generating multi-view images from a single-view input is an essential yet challenging problem. It has broad applications in vision, graphics, and robotics. Our study indicates that the widely-used generative adversarial network (GAN) may learn ``incomplete'' representations due to the single-pathway framework: an encoder-decoder network followed by a discriminator network. We propose CR-GAN to address this problem. In addition to the single reconstruction path, we introduce a generation sideway to maintain the completeness of the learned embedding space. The two learning pathways collaborate and compete in a parameter-sharing manner, yielding considerably improved generalization ability to ``unseen'' dataset. More importantly, the two-pathway framework makes it possible to combine both labeled and unlabeled data for self-supervised learning, which further enriches the embedding space for realistic generations. 
The experimental results prove that CR-GAN significantly outperforms state-of-the-art methods, especially when generating from ``unseen'' inputs in wild conditions. \footnote{ The code and pre-trained models are publicly available: \url{https://github.com/bluer555/CR-GAN}}

\end{abstract}

\section{Introduction}
Generating multi-view images from a single-view input is an interesting problem with broad applications in vision, graphics, and robotics. Yet, it is a challenging problem since 1) computers need to ``imagine'' what a given object would look like after a 3D rotation is applied; and 2) the multi-view generations should preserve the same ``identity''.

Generally speaking, previous solutions to this problem include model-driven synthesis \cite{blanz1999morphable}, data-driven generation \cite{zhu2014multi,yan2016perspective}, and a combination of the both \cite{zhuxiangyu2016,rezende2016unsupervised}. Recently, generative adversarial networks (GANs) \cite{goodfellow2014generative} have shown impressive results in multi-view generation \cite{tran2017disentangled,zhao2017multi}.

\begin{figure}[t]
\centering
\includegraphics[width=0.96\linewidth]{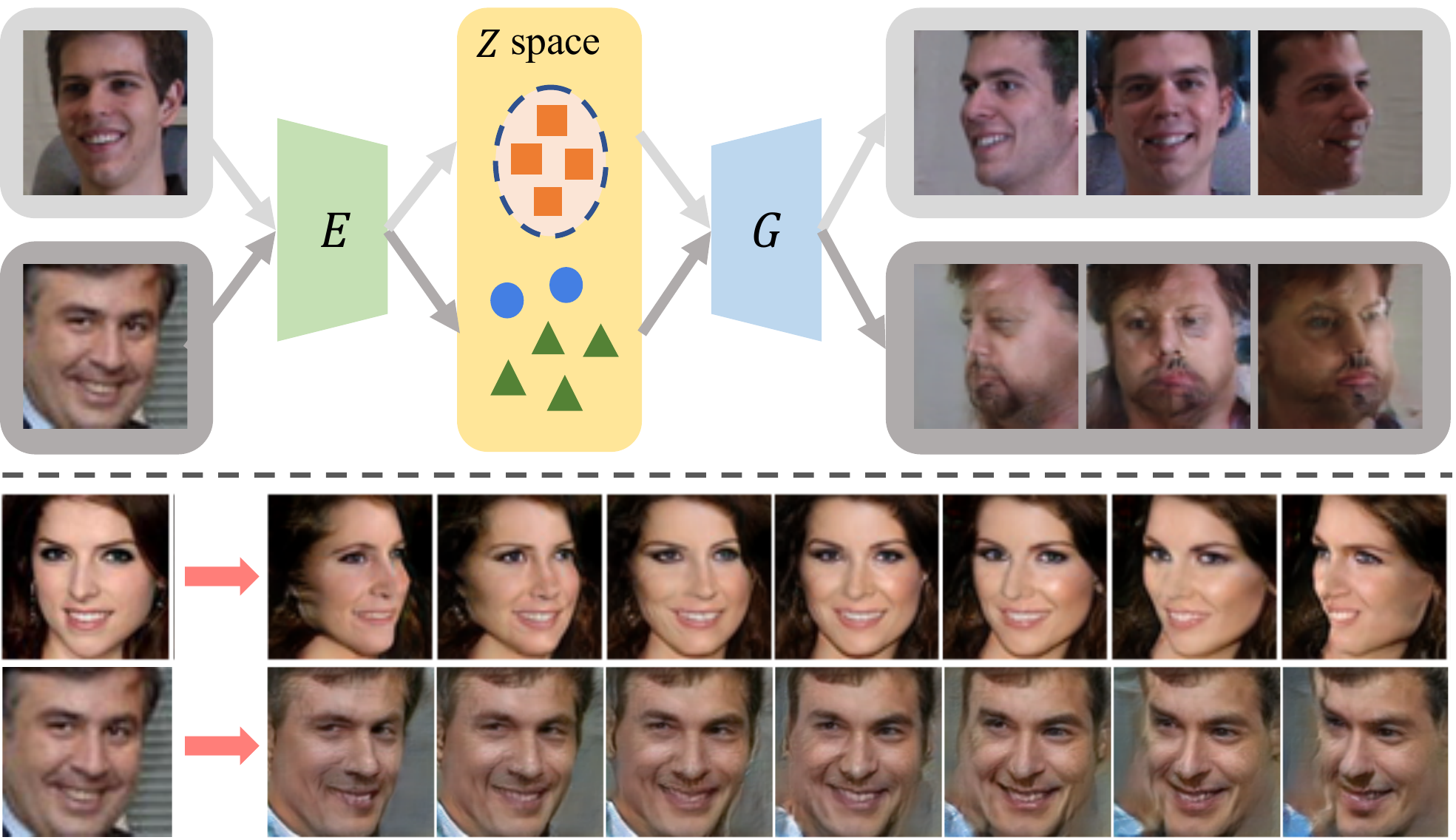}
\caption{Top: The limitation of existing GAN-based methods. They can generate good results if the input is mapped into the learned subspace (Row 1). However, ``unseen'' data may be mapped out of the subspace, leading to poor results (Row 2). Bottom: Our results. By learning complete representations, the proposed CR-GAN can generate realistic, identity-preserved images from a single-view input.}
\label{fig:fig_intro}
\vspace{-4mm}
\end{figure}

These GAN-based methods usually have a single-pathway design: an encoder-decoder network is followed by a discriminator network. The encoder ($E$) maps input images into a latent space ($Z$), where the embeddings are first manipulated and then fed into the decoder ($G$) to generate novel views. 

However, our experiments indicate that this single-pathway design may have a severe issue: they can only learn ``incomplete'' representations, yielding limited generalization ability on ``unseen'' or unconstrained data. Take Fig. \ref{fig:fig_intro} as an example. During the training, the outputs of $E$ constitute only a subspace of $Z$ since we usually have a limited number of training samples. This would make $G$ only ``see'' part of $Z$. During the testing, it is highly possible that $E$ would map an ``unseen'' input outside the subspace. As a result, $G$ may produce poor results due to the unexpected embedding.

To address this issue, we propose CR-GAN to learn \textit{Complete Representations} for multi-view generation. The main idea is, in addition to the reconstruction path, we introduce another generation path to create view-specific images from embeddings that are randomly sampled from $Z$. Please refer to Fig. \ref{fig:our_model} for an illustration. The two paths share the same $G$. In other words, $G$ learned in the generation path will guide the learning of both $E$ and $D$ in the reconstruction path, and vice versa. $E$ is forced to be an inverse of $G$, yielding complete representations that would span the entire $Z$ space. More importantly, the two-pathway learning can easily utilize both labeled and unlabeled data for self-supervised learning, which can largely enrich the $Z$ space for natural generations. 

To sum up, we have following contributions:
\begin{itemize}
\item To the best of our knowledge, we are the first to investigate ``complete representations'' of GAN models;

\item We propose CR-GAN that can learn ``complete'' representations, using a two-pathway learning scheme;

\item CR-GAN can leverage unlabeled data for self-supervised learning, yielding improved generation quality;

\item CR-GAN can generate high-quality multi-view images from even ``unseen'' dataset in wild conditions.
\end{itemize}

\section{Related Work}

\textbf{Generative Adversarial Networks (GANs)}. 
Goodfellow~\etal~\cite{goodfellow2014generative} introduced GAN to estimate target distribution via an adversarial process. 
Gulrajani~\etal~\cite{gulrajani2017improved} presented a more stable approach to enforce \textit{Lipschitz Constraint} on Wasserstein GAN~\cite{arjovsky2017wasserstein}.
AC-GAN~\etal~\cite{odena2016conditional} extended the discriminator by containing an auxiliary decoder network to estimate class labels for the training data. BiGANs~\cite{donahue2016adversarial,dumoulin2016adversarially} try to learn an inverse mapping to project data back into the latent space. Our method can also find an inverse mapping, make a balanced minimax game when training data is limited.

\textbf{Multi-view Synthesis}.
Hinton~\etal~\cite{hinton2011transforming} introduced transforming auto-encoder to generate images with view variance. Yan~\etal~\cite{yan2016perspective} proposed Perspective Transformer Nets to find the projection transformation. Zhou~\etal~\cite{zhou2016view} propose to synthesize views by appearance flow. Very recently, GAN-based methods usually follow a single-pathway design: an encoder-decoder network~\cite{peng2016recurrent} followed by a discriminator network. For example, to normalize the viewpoint, {\it e.g.} face frontalization, they either combine encoder-decoder with 3DMM~\cite{blanz1999morphable} parameters~\cite{yin2017towards}, or use duplicates to predict global and local details~\cite{huang2017beyond}. DR-GAN~\cite{tran2017disentangled} follows the single-pathway framework to learn identity features that are invariant to viewpoints. However, it may learn ``incomplete'' representations due to the single-pathway framework. In contrast, CR-GAN can learn complete representations using a two-pathway network, which guarantees high-quality generations even for ``unseen'' inputs.

\textbf{Pose-Invariant Representation Learning}. 
For representation learning~\cite{li2016multi,fan2016learning}, early works may use \textit{Canonical Correlation Analysis} to analyze the commonality among different pose subspaces~\cite{hardoon2004canonical,peng2015circle}. Recently, deep learning based methods use synthesized images to disentangle pose and identity factors by cross-reconstruction ~\cite{zhu2014multi,peng2017recons}, or transfer information from pose variant inputs to a frontalized appearance~\cite{zhu2013deep}. However, they usually use only labeled data, leading to a limited performance. We proposed a two-pathway network to leverage both labeled and unlabeled data for self-supervised learning, which can generate realistic images in challenging conditions.
\section{Proposed Method} \label{sec:method}

\subsection{A Toy Example of Incomplete Representations} \label{sec:toy}
\begin{figure}[t]
	\centering
	\includegraphics[width=0.9\linewidth]{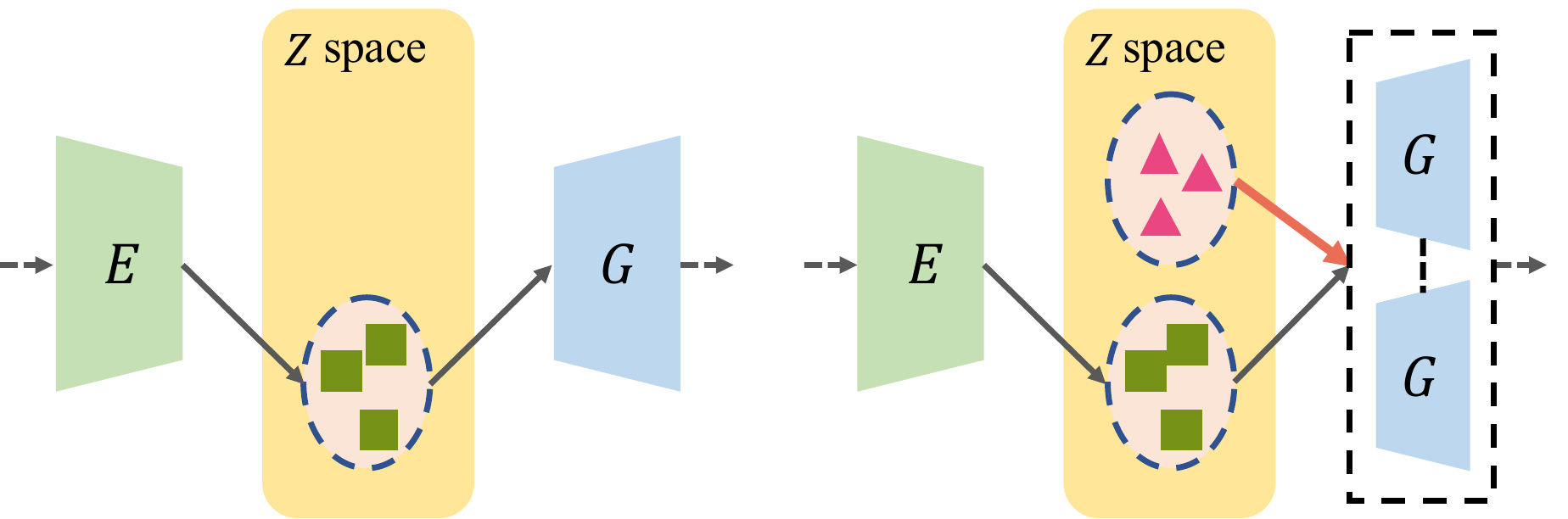}
	\caption{Left: Previous methods use a single path to learn the latent representation, but it is incomplete in the whole space. Right: We propose a two-pathway network combined with self-supervised learning, which can learn complete representations. }
	\label{fig:our_model}
	\vspace{-4mm}
\end{figure}

A single-pathway network, {\it i.e.} an encoder-decoder network followed by a discriminator network, may have the issue of learning ``incomplete'' representations. As illustrated in Fig.~\ref{fig:our_model} left, the encoder $E$ and decoder $G$ can ``touch'' only a subspace of $Z$ since we usually have a limited number of training data. This would lead to a severe issue in testing when using ``unseen'' data as the input. It is highly possible that $E$ may map the novel input out of the subspace, which inevitably leads to poor generations since $G$ has never ``seen'' the embedding.

A toy example is used to explain this point. We use Multi-PIE~\cite{Gross2010MultiPIE} to train a single-pathway network. As shown in the top of Fig.~\ref{fig:fig_intro}, the network can generate realistic results on Multi-PIE (the first row), as long as the input image is mapped into the learned subspace. However, when testing ``unseen'' images from IJB-A~\cite{klare2015pushing}, the network may produce unsatisfactory results (the second row). In this case, the new image is mapped out of the learned subspace.

This fact motivates us to train $E$ and $G$ that can ``cover'' the whole $\mathit{Z}$ space, so we can learn complete representations. We achieve this goal by introducing a separate generation path, where the generator focuses on mapping the entire $\mathit{Z}$ space to high-quality images. Fig.~\ref{fig:our_model} illustrates the comparison between the single-pathway and two-pathway networks. Please refer to Fig.~\ref{fig:fig_framework} (d) for an overview of our approach.
\vspace{-1mm}

\subsection{Generation Path} \label{sec:generation-path}

The generation path trains generator $G$ and discriminator $D$. Here the encoder $E$ is not involved since $G$ tries to generate from random noise. Given a view label $v$ and random noise $\mathbf{z}$, $G$ aims to produce a realistic image $G(v,\mathbf{z})$ under view $v$. $D$ is trying to distinguish real data from $G$'s output, which minimizes:
\begin{equation}
\label{WACGAN-gp_D}
\small
\begin{gathered}
\underset{\mathbf{z}\sim\mathbb{P}_{\mathbf{z}}}{\mathbb{E}}[D_s(G(v, \mathbf{z}))] - \underset{\mathbf{x}\sim\mathbb{P}_{\mathbf{x}}}{\mathbb{E}}[D_s(\mathbf{x})] +
\\
\lambda_1 \underset{\hat{\mathbf{x}}\sim\mathbb{P}_{\hat{{\mathbf{x}}}}}{\mathbb{E}}[(\left\| \bigtriangledown_{\hat{\mathbf{x}}}D(\hat{\mathbf{x}}) \right \|_2 - 1)^2] - \lambda_2\underset{\mathbf{x}\sim\mathbb{P}_{\mathbf{x}}}{\mathbb{E}}[P(D_v(\mathbf{x})=v)],
\end{gathered}
\end{equation}
where $\mathbb{P}_{\mathbf{x}}$ is the data distribution and $\mathbb{P}_{\mathbf{z}}$ is the noise uniform distribution, $\mathbb{P}_{\hat{\mathbf{x}}}$ is an interpolation between pairs of points sampled from data distribution and the generator distribution~\cite{gulrajani2017improved}. $G$ tries to fool $D$, it maximizes:
\begin{equation}
\label{WACGAN-gp_G}
\small
\underset{\mathbf{z}\sim\mathbb{P}_{\mathbf{z}}}{\mathbb{E}}[D_s(G(v, \mathbf{z}))] + \lambda_3\underset{\mathbf{z}\sim\mathbb{P}_{\mathbf{z}}}{\mathbb{E}}[P(D_v({G(v, \mathbf{z})})=v)],
\end{equation}
where $(D_v(\cdot), D_s(\cdot)) = D(\cdot)$ denotes pairwise outputs of the discriminator. $D_v(\cdot)$ estimates the probability of being a specific view, $D_s(\cdot)$ describes the image quality, \ie, how real the image is. Note that in Eq.~\ref{WACGAN-gp_D}, $D$ learns how to estimate the correct view of a real image~\cite{odena2016conditional}, while $G$ tries to produce an image with that view in order to get a high score from $D$ in Eq.~\ref{WACGAN-gp_G}.

\begin{algorithm}[t]
    \KwIn{Sets of view labeled images $\mathit{X}$, max number of steps $T$, and batch size $m$.}
    \KwOut{Trained network $E$, $G$ and $D$.}
    \For{$t = 1$ \KwTo $T$}{
        \For{$i = 1$ \KwTo $m$}{
            1. Sample $\mathbf{z}\sim\mathit{P}_{\mathbf{z}}$ and $\mathbf{x}_i \sim \mathit{P}_{\mathbf{x}}$ with $v_i$\;
            2. $\bar{\mathbf{x}} \gets G(v_i, \mathbf{z})$\;
            3. Update $D$ by Eq.~\ref{WACGAN-gp_D}, and $G$ by Eq.~\ref{WACGAN-gp_G}\;
            4. Sample $\mathbf{x}_j \sim \mathit{P}_{\mathbf{x}}$ with $v_j$ (where $\mathbf{x}_j$ and $\mathbf{x}_i$ share the same identity)\;
            5. $(\bar{\mathbf{v}}, \bar{\mathbf{z}}) \gets E(\mathbf{x}_i)$\;
            6. $\tilde{\mathbf{x}}_j \gets G(v_j, \bar{\mathbf{z}})$\;
            7. Update $D$ by Eq.~\ref{Reconstruction_D}, and $E$ by Eq.~\ref{Reconstruction_E}\;
        }
    }
    \caption{Supervised training with two paths}\label{xxx-GAN}
\end{algorithm}
\vspace{-1mm}

\subsection{Reconstruction Path}\label{sec:reconstruction-path}

The reconstruction path trains $E$ and $D$ but keeping $G$ fixed. $E$ tries to reconstruct training samples, this would guarantee that $E$ will be learned as an inverse of $G$, yielding complete representations in the latent embedding space.

The output of $E$ should be identity-preserved so the multi-view images will present the same identity. We propose a cross reconstruction task to make $E$ disentangle the viewpoint from the identity. More specifically, we sample a real image pair $(\mathbf{x}_i, \mathbf{x}_j)$ that share the same identity but different views $v_i$ and $v_j$. The goal is to reconstruct $x_j$ from $x_i$. To achieve this, $E$ takes $\mathbf{x}_i$ as input and outputs an identity-preserved representation $\bar{\mathbf{z}}$ together with the view estimation $\bar{\mathbf{v}}$: $ (\bar{\mathbf{v}}, \bar{\mathbf{z}}) = (E_v(\mathbf{x}_i), E_z(\mathbf{x}_i)) = E(\mathbf{x}_i)$. Note that $\bar{\mathbf{v}}$ is learned for further self-supervised training as shown in Sec.~\ref{sec:unlabeled-dataset}.

$G$ takes $\bar{\mathbf{z}}$ and view $v_j$ as the input. As $\bar{\mathbf{z}}$ is expected to carry the identity information of this person, with view $v_j$'s help, $G$ should produce $\tilde{\mathbf{x}}_j$, the reconstruction of $\mathbf{x}_j$. $D$ is trained to distinguish the fake image $\tilde{\mathbf{x}}_j$ from the real one $\mathbf{x}_i$. Thus $D$ minimizes:
\begin{equation}
\small
\begin{gathered}
\label{Reconstruction_D}
\underset{\mathbf{x}_i, \mathbf{x}_j\sim\mathbb{P}_{\mathbf{x}}}{\mathbb{E}}[D_s(\tilde{\mathbf{x}}_j) - D_s(\mathbf{x}_i)] + 
\\
\lambda_1 \underset{\hat{\mathbf{x}}\sim\mathbb{P}_{\hat{\mathbf{x}}}}{\mathbb{E}}[(\left\| \bigtriangledown_{\hat{\mathbf{x}}}D(\hat{\mathbf{x}}) \right \|_2 - 1)^2] - \lambda_2\underset{\mathbf{x}_i\sim\mathbb{P}_x}{\mathbb{E}}[P(D_v(\mathbf{x}_i)=v_i)],
\end{gathered}
\end{equation}
where $\tilde{\mathbf{x}}_j = G(v_j, E_z(\mathbf{x}_i))$. $E$ helps $G$ to generate high quality image with view $v_j$, so $E$ maximizes:
\begin{equation}
\small
\begin{gathered}
\label{Reconstruction_E}
\underset{\mathbf{x}_i, \mathbf{x}_j\sim\mathbb{P}_{\mathbf{x}}}{\mathbb{E}}[D_s(\tilde{\mathbf{x}}_j) + \lambda_3 P(D_v(\tilde{\mathbf{x}}_j)=v_j) - 
\\
\lambda_4 L_1(\tilde{\mathbf{x}}_j, \mathbf{x}_j) - \lambda_5 L_v(E_v(\mathbf{x}_i), v_i)],
\end{gathered}
\end{equation}
where $L_1$ loss is utilized to enforce that $\tilde{x}_j$ is the reconstruction of $x_j$. $L_v$ is the cross-entropy loss of estimated and ground truth views, to let $E$ be a good view estimator.

The two-pathway network learns complete representations: First, in the generation path, $G$ learns how to produce real images from \textit{any} inputs in the latent space. Then, in the reconstruction path, $G$ retains the generative ability since it keeps unchanged. The alternative training details of the two pathways are summarized in Algorithm~\ref{xxx-GAN}.

\begin{algorithm}[t]
    \KwIn{Sets of view labeled and unlabeled images $\mathit{X}$, max number of steps $T$, and batch size $m$.}
    \KwOut{Trained network $E$, $G$ and $D$.}
    Pre-train $E$, $G$ and $D$ according to Algorithm~\ref{xxx-GAN}\;
    \For{$t = 1$ \KwTo $T$}{
        \For{$i = 1$ \KwTo $m$}{
            Sample $\mathbf{z}\sim\mathit{P}_{\mathbf{z}}$ and $\mathbf{x} \sim \mathit{P}_{\mathbf{x}}$\;
            \eIf{$\mathbf{x}$ is labeled}{
                1. $\mathbf{x}_i \gets \mathbf{x}$\;
                2. Get the label $v_i$ of $\mathbf{x}_i$\;
                3. Repeat the step 2 to 7 in Algorithm~\ref{xxx-GAN}\;
            }{
                4. $(\bar{\mathbf{v}}, \bar{\mathbf{z}}) \gets E(\mathbf{x})$\;
                5. Compute $\hat{v}$ (the estimation of $\bar{\mathbf{v}}$)\;
                6. Update $D$ by Eq.~\ref{self_supervise_recon_D} and $E$ by Eq.~\ref{self_supervise_recon_E}\;
                7. Update $D$ by Eq.~\ref{self_supervise_gen_D} and $G$ by Eq.~\ref{self-supervise_gen_G}\;
            }
        }
    }
    \caption{Self-supervised training with two paths}\label{xxx-GAN_unlabeled}
\end{algorithm}

\begin{figure*}[t]
	\centering
	\includegraphics[width=0.96\linewidth]{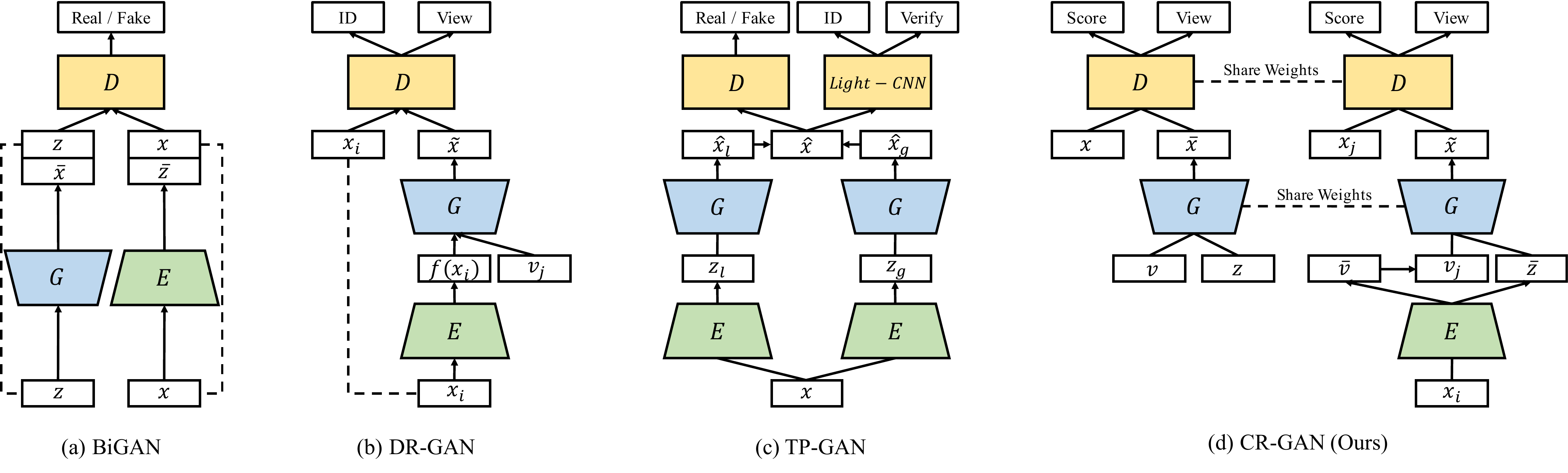}
	\caption{Comparison of BiGAN~\protect\cite{donahue2016adversarial}, DR-GAN~\protect\cite{tran2017disentangled}, TP-GAN~\protect\cite{huang2017beyond} and our method.}
	\label{fig:fig_framework}
	\vspace{-4mm}
\end{figure*}
\vspace{-1mm}

\subsection{Self-supervised Learning}\label{sec:unlabeled-dataset}

Labeled datasets are usually limited and biased. For example, Multi-PIE~\cite{Gross2010MultiPIE} is collected in a constrained setting, while large-pose images in 300wLP~\cite{zhuxiangyu2016} are distorted. As a result, $G$ would generate low-quality images since it has only ``seen'' poor and limited examples.  

To solve this issue, we further improve the proposed CR-GAN with self-supervised learning. The key idea is to use a pre-trained model to estimate viewpoints for unlabeled images. Accordingly, we modify the supervised training algorithm into two phases. In the first stage, we pre-train the network on labeled data to let $E$ be a good view estimator. In the second stage, both labeled and unlabeled data are utilized to boost $G$. When an unlabeled image $\mathbf{x}$ is fed to the network, a view estimation $\bar{\mathbf{v}}$ is obtained by $E_v(\cdot)$. Denote $\hat{v}$ to be the closest one-hot vector of $\bar{\mathbf{v}}$, in the reconstruction path, we let $E$ minimize $L_v(\bar{\mathbf{v}}, \hat{v})$ and then reconstruct $\mathbf{x}$ to itself. Similar to Eq.~\ref{Reconstruction_D}, $D$ minimizes:
\begin{equation}
\small
\begin{gathered}
\label{self_supervise_recon_D}
\underset{\mathbf{x}\sim\mathbb{P}_{\mathbf{x}}}{\mathbb{E}}[D_s(G(\hat{v}, E_z(\mathbf{x}))) - D_s(\mathbf{x})] + 
\\
\lambda_1 \underset{\hat{\mathbf{x}}\sim\mathbb{P}_{\hat{\mathbf{x}}}}{\mathbb{E}}[(\left\| \bigtriangledown_{\hat{\mathbf{x}}}D(\hat{\mathbf{x}}) \right \|_2 - 1)^2] - \lambda_2\underset{\mathbf{x}\sim\mathbb{P}_{\mathbf{x}}}{\mathbb{E}}[P(D_v(\mathbf{x})=\hat{v})],
\end{gathered}
\end{equation}
similar to Eq.~\ref{Reconstruction_E}, $E$ maximizes:
\begin{equation}
\small
\begin{gathered}
\label{self_supervise_recon_E}
\underset{\mathbf{x}\sim\mathbb{P}_{\mathbf{x}}}{\mathbb{E}}[D_s(G(\hat{v}, E_z(\mathbf{x}))) + \lambda_3 P(D_v({G(\hat{v}, E_z(\mathbf{x}))})=\hat{v}) - 
\\
\lambda_4 L_1(G(\hat{v}, E_z(\mathbf{x})), \mathbf{x}) - \lambda_5 L_v(E_v(\mathbf{x}), \hat{v})].
\end{gathered}
\end{equation}
In the generation path, we let $\hat{v}$ be the ground truth of $\mathbf{x}$, and generate an image in view $\hat{v}$. So similar to Eq.~\ref{WACGAN-gp_D}, $D$ minimizes:
\begin{equation}
\small
\begin{gathered}
\label{self_supervise_gen_D}
\underset{\mathbf{z}\sim\mathbb{P}_{\mathbf{z}}}{\mathbb{E}}[D_s(G(\hat{v}, \mathbf{z}))] - \underset{\mathbf{x}\sim\mathbb{P}_{\mathbf{x}}}{\mathbb{E}}[D_s(\mathbf{x})] + 
\\
\lambda_1 \underset{\hat{\mathbf{x}}\sim\mathbb{P}_{\hat{\mathbf{x}}}}{\mathbb{E}}[(\left\| \bigtriangledown_{\hat{\mathbf{x}}}D(\hat{\mathbf{x}}) \right \|_2 - 1)^2] - \lambda_2\underset{\mathbf{x}\sim\mathbb{P}_{\mathbf{x}}}{\mathbb{E}}[P(D_v(\mathbf{x})=\hat{v})],
\end{gathered}
\end{equation}
similar to Eq.~\ref{WACGAN-gp_G}, $G$ maximizes:
\begin{equation}
\small
\label{self-supervise_gen_G}
\underset{\mathbf{z}\sim\mathbb{P}_{\mathbf{z}}}{\mathbb{E}}[D_s(G(\hat{v}, \mathbf{z}))] + \lambda_3\underset{\mathbf{z}\sim\mathbb{P}_{\mathbf{z}}}{\mathbb{E}}[P(D_v({G(\mathbf{z})})=\hat{v})].
\end{equation}

Once we get the pre-trained model, the encoder $E$ predicts the probabilities of the input image belonging to different views. We choose the view with the highest probability as the estimation. Our strategy is similar to RANSAC algorithm~\cite{fischler1987random}, where we treat the estimations with higher confidence as ``inliers'' and use them to make view estimation more accurate. We summarize the self-supervised training in Algorithm~\ref{xxx-GAN_unlabeled}.

Compared with the single-pathway solution, the proposed two-pathway network boosts the self-supervised learning in two aspects: 1) it provides a better pre-trained model for viewpoint estimation as a byproduct; and 2) it guarantees that we can take full advantage of unlabeled data in training since CR-GAN learns complete representations.
\vspace{-1mm}

\subsection{Discussion}\label{sec:comparisons}

To highlight the novelty of our method, we compare CR-GAN with the following three GANs. In Fig.~\ref{fig:fig_framework}, we show their network structures as well as ours for visual comparison.
 
\textbf{BiGAN}~\cite{donahue2016adversarial,dumoulin2016adversarially} jointly learns a generation network $G$ and an inference network $E$. The authors proved that $E$ is an inverse network of $G$.
However, in practice, BiGAN produces poor reconstructions due to finite data and limited network capacity. Instead, CR-GAN uses explicit reconstruction loss to solve this issue.

\textbf{DR-GAN}~\cite{tran2017disentangled} also tries to learn an identity preserved representation to synthesize multi-view images. But we have two distinct differences. First, the output of its encoder, also acts as the decoder's input, completely depends on the training dataset. Therefore, it can not deal with new data. Instead, we use the generation path to make sure that the learning of our $G$ is ``complete''. Second, we don't let $D$ estimate the identity for training data, because we employ unlabeled dataset in self-supervised learning which has no identity information. The involvement of unlabeled dataset also makes our model more robust for ``unseen'' data.

\textbf{TP-GAN}~\cite{huang2017beyond} uses two pathway GANs for frontal view synthesis. Their framework is different from ours: First, they use two distinct encoder-decoder networks, while CR-GAN shares all modules in the two pathways. Besides, they use two pathways to capture global features and local details, while we focus on learning complete representations in multi-view generation.
\begin{figure*}[t]
	\centering
	\includegraphics[width=\linewidth]{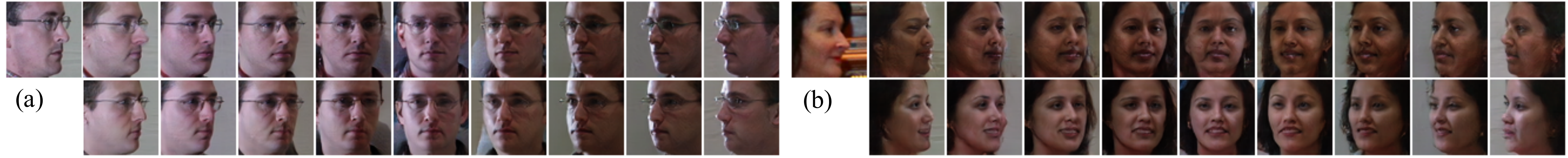}
	\caption{Results generated by the single-pathway and two-pathway from (a) Multi-PIE~\protect\cite{Gross2010MultiPIE} and (b) IJB-A~\protect\cite{klare2015pushing}. In each case, the images generated by the two-pathway (Row 2) outperform the ones produced by the single-pathway (Row 1).}
	\label{fig:single-two-rotation}
	\vspace{-4mm}
\end{figure*}

\begin{figure*}[t]
	\centering
	\includegraphics[width=\linewidth]{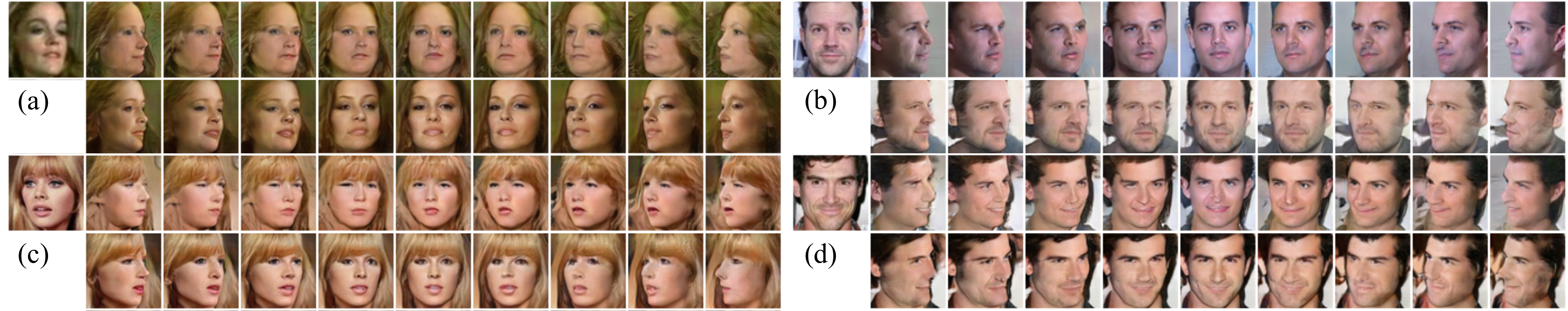}
	\caption{Multi-view face generation results on CelebA~\protect\cite{liu2015deep}. In each case, self-supervised learning (Row 2) generates more realistic images than supervised learning (Row 1). Note that in (b) and (d), the beard and eyebrows are well-kept in different views.}
	\label{fig:celeba}
	\vspace{-4mm}
\end{figure*}

\section{Experiments}\label{sec:experiments}

CR-GAN aims to learn complete representations in the embedding space. We achieve this goal by combining the two-pathway architecture with self-supervised learning. We conduct experiments to evaluate these two contributions respectively. Then we compare our CR-GAN with DR-GAN~\cite{tran2017disentangled}, both the visual results and t-SNE visualization in the embedding space are shown. We also compare CR-GAN and BiGAN with an image reconstruction task.

\subsection{Experimental Settings}

\textbf{Datasets}. We evaluate CR-GAN on datasets with and without view labels. Multi-PIE~\cite{Gross2010MultiPIE} is a labeled dataset collected under constrained environment. We use 250 subjects from the first session with 9 poses within $\pm 60^{\circ}$, 20 illuminations, and two expressions. The first 200 subjects are for training and the rest 50 for testing. 300wLP~\cite{zhuxiangyu2016} is augmented from 300W~\cite{sagonas2013300} by the face profiling approach~\cite{zhuxiangyu2016}, which contains view labels as well. We employ images with yaw angles ranging from $-60^{\circ}$ to $+60^{\circ}$, and discretize them into 9 intervals.

For evaluation on unlabeled datasets, we use CelebA~\cite{liu2015deep} and IJB-A~\cite{klare2015pushing}. CelebA contains a large amount of celebrity images with unbalanced viewpoint distributions. Thus, we collect a subset of 72,000 images from it, which uniformly ranging from $-60^{\circ}$ to $+60^{\circ}$. Notice that the view labels of the images in CelebA are only utilized to collect the subset, while no view or identity labels are employed in the training process. We also use IJB-A which contains 5,396 images for evaluation. This dataset is challenging, since there are extensive identity and pose variations.

\textbf{Implementation Details}. Our network implementation is modified from the residual networks in WGAN-GP~\cite{gulrajani2017improved}, where $E$ shares a similar network structure with $D$. During training, we set $v$ to be a one-hot vector with 9 dimensions and $z \in [-1, 1]^{119}$ in the latent space. The batch size is 64. Adam optimizer~\cite{kingma2014adam} is used with the learning rate of $0.0005$ and momentum of $[0, 0.9]$. According to the setting of WGAN-GP, we let $\lambda_1 = 10$, $\lambda_2 \sim \lambda_4 = 1$, $\lambda_5 = 0.01$. Moreover, all the networks are trained after 25 epochs in supervised learning; we train 10 more epochs in self-supervised learning.

\subsection{Single-pathway {\it vs.} Two-pathway}\label{sec-single-two}

We compare two-pathway network with the one using a single reconstruction path. All the networks are trained on Multi-PIE. When test with Muli-PIE, as shown in Fig.~\ref{fig:single-two-rotation}~(a), both models produce desirable results. In each view, facial attributes like glasses are kept well. However, single-pathway model gets unsatisfactory results on IJB-A, which is an ``unseen'' dataset. As shown in Fig.~\ref{fig:single-two-rotation}~(b), two-pathway model consistently produce natural images with more details and fewer artifacts. Instead, the single-pathway model cannot generate images with good quality. This result prove that our two-pathway network handles ``unseen'' data well by learning a complete representation in the embedding space.

\subsection{Supervised {\it vs.} Self-supervised Learning}\label{sec:sup-self-sup}

The two-pathway network is employed in the following evaluations. We use Multi-PIE and 300wLP in supervised learning. For self-supervised learning, in addition to the above datasets, CelebA is employed as well. Note that we don't use view or identity labels in CelebA during training.

\begin{figure*}[t]
	\centering
	\includegraphics[width=\linewidth]{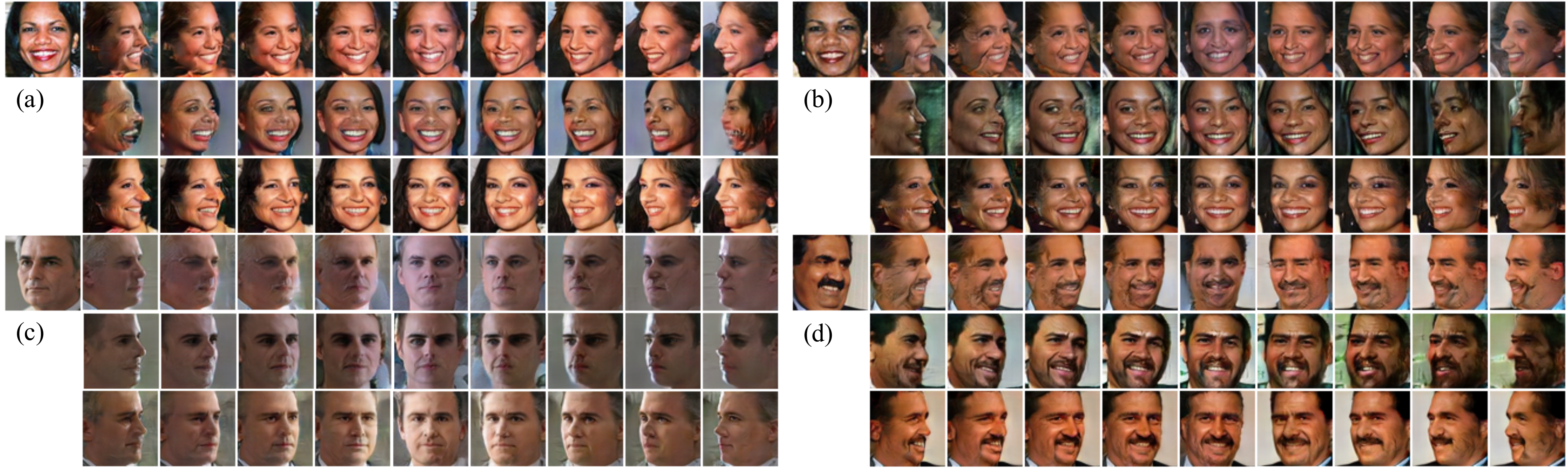}
	\caption{Multi-view face generation results on IJB-A~\protect\cite{klare2015pushing}. In each case, from top to bottom: the results generated by our supervised model, DR-GAN~\protect\cite{tran2017disentangled} and our self-supervised model. DR-GAN fails to produce favourable images of large poses, while our method can synthesize reasonable profile images.}
	\label{fig:IJBA}
	\vspace{-4mm}
\end{figure*}

\textbf{Evaluation on CelebA}. Fig.~\ref{fig:celeba} shows the results on CelebA. In Fig.~\ref{fig:celeba} (a), although the supervised model generates favorable results, there are artifacts in all views. As the supervised model is only trained on Multi-PIE and 300wLP, it is difficult to ``approximate'' the data in the wild. Instead, the self-supervised model has learned a latent representation where richer features are embedded, so it generates more realistic results while the identities are well preserved. We can make the similar observation in Fig.~\ref{fig:celeba} (b). The supervised model can only generate images that are similar to Multi-PIE, while the self-supervised model can generate novel identities. In Fig.~\ref{fig:celeba} (c) and (d), the self-supervised model preserve identity and attributes in a better way than others.

\textbf{Evaluation on IJB-A}. Fig.~\ref{fig:IJBA} shows more results on IJB-A. We find that our self-supervised model successfully generalize what it has learned from CelebA to IJB-A. Note that it is our self-supervised learning approach that makes it possible to train the network on unlabeled datasets.

\subsection{Comparison with DR-GAN}\label{sec:unlabeled}

Furthermore, we compare our self-supervised CR-GAN with DR-GAN~\cite{tran2017disentangled}. We replace DC-GAN~\cite{radford2015unsupervised} network architecture used in DR-GAN with WGAN-GP for a fair comparison.

\textbf{Evaluation on IJB-A}. We show the results of DR-GAN and CR-GAN in Fig.~\ref{fig:IJBA} respectively. DR-GAN produces sharp images, but the facial identities are not well-kept. By contrast, in Fig.~\ref{fig:IJBA} (a) and (b), CR-GAN produces face images with similar identities. In all cases, DR-GAN fails to produce high-quality images with large poses. Although not perfect enough, CR-GAN can synthesize reasonable profile images.

\begin{figure}[t]
	\centering
	\includegraphics[width=\linewidth]{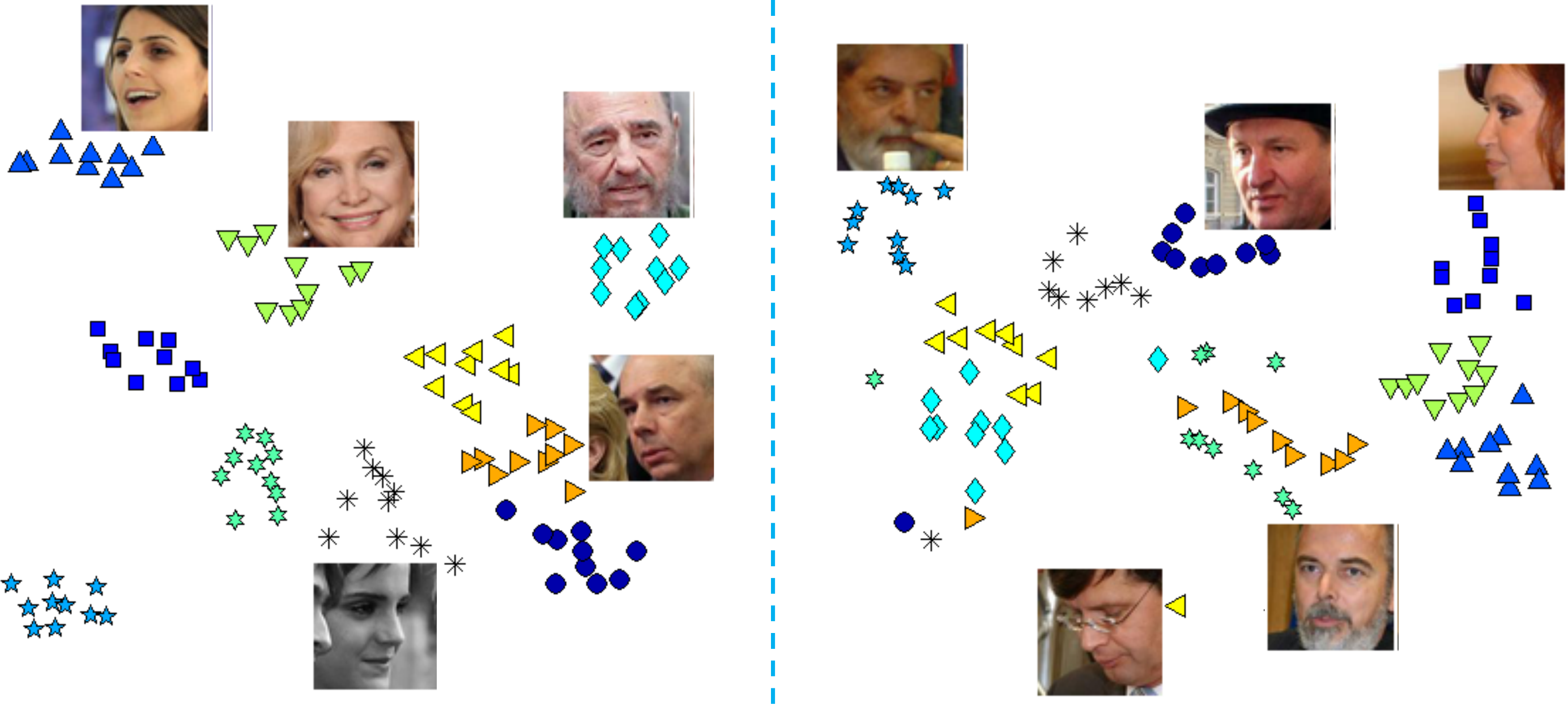}
	\caption{t-SNE visualization for the embedding space of CR-GAN (\textbf{left}) and DR-GAN (\textbf{right}), with 10 subjects from IJB-A~\protect\cite{klare2015pushing}. The same marker shape (color) indicates the same subject. For CR-GAN, multi-view images of the same subject are embedded close to each other, which means the identities are better preserved.}
	\label{fig:t-SNE}
	\vspace{-1mm}
\end{figure}

\textbf{Identity Similarities on IJB-A}. We generate 9 views for each image in IJB-A both using DR-GAN and CR-GAN. Then we obtain a 128-dim feature for each view by FaceNet~\cite{SchroffCVPR15}. We evaluate the identity similarities between the real and generated images by feeding them to FaceNet. The squared L2 distances of the features directly corresponding to the face similarity: faces of the same subjects have small distances, while faces of different subjects have large distances. Table~\ref{tb:facenet} shows the results of the average L2 distance of CR-GAN and DR-GAN in different datasets. Our method outperforms DR-GAN on all datasets, especially on IJB-A which contains unseen data. Fig.~\ref{fig:t-SNE} shows the t-SNE visualization in the embedding space of DR-GAN and CR-GAN respectively. For clarity, we only visualize 10 randomly selected subjects along with 9 generated views of each. Compared with DR-GAN, CR-GAN produces tighter clusterings: multi-view images of the same subject are embedded close to each other. It means the identities are better preserved.

\begin{table}[t]
\begin{center}
\setlength\tabcolsep{2.0pt}
\begin{tabular}{c|ccc}
\hline
 & Multi-PIE & CelebA & IJB-A \\
\hline
DR-GAN & $1.073 \pm 0.013$ & $1.281 \pm 0.007$ & $1.295 \pm 0.008$ \\
CR-GAN & $\textbf{1.018} \pm \textbf{0.019}$ & $\textbf{1.214} \pm \textbf{0.009}$ & $\textbf{1.217} \pm \textbf{0.010}$ \\
\hline
\end{tabular}
\end{center}
\vspace{-3mm}
\caption{Identity similarities between real and generated images.}
\label{tb:facenet}
\vspace{-3mm}
\end{table}

\begin{figure}[t]
	\centering
	\includegraphics[width=\linewidth]{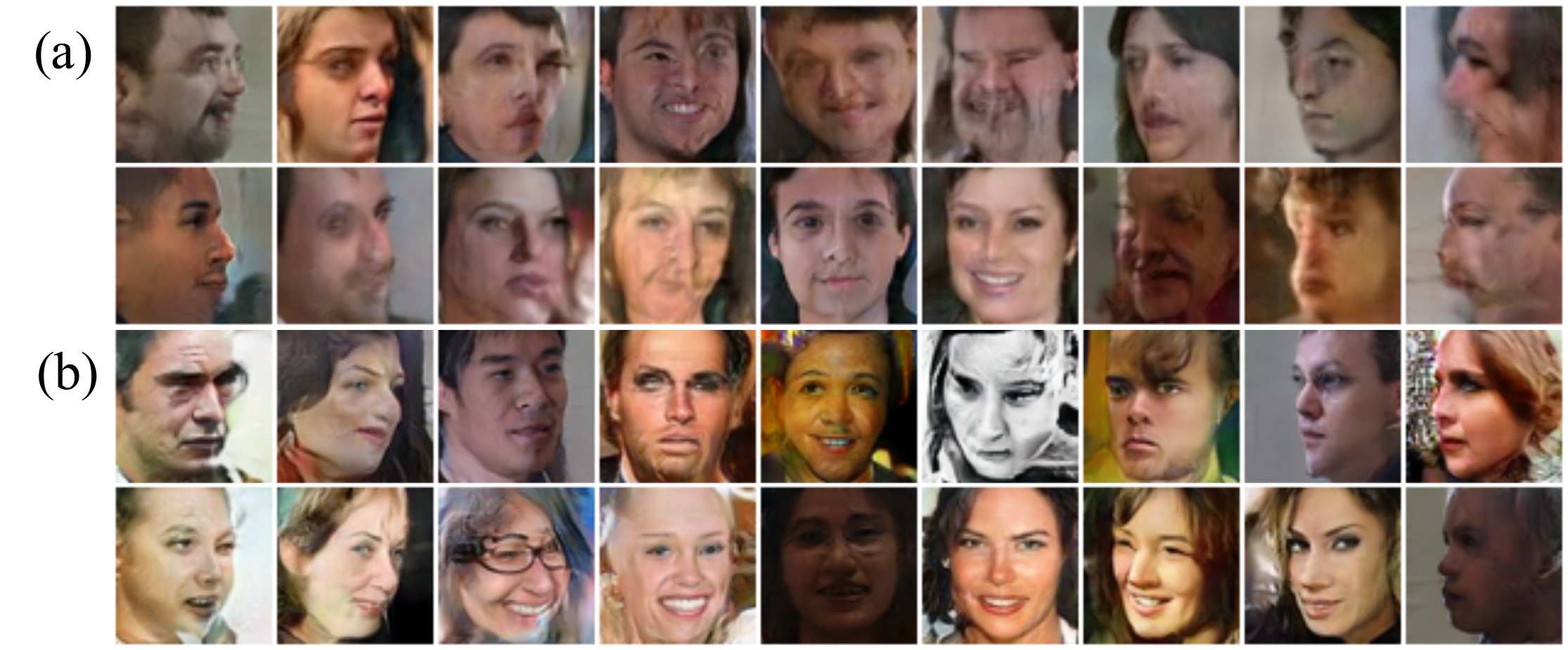}
	\caption{Generating multi-view images from the random noise. (a) DR-GAN~\protect\cite{tran2017disentangled} generates blurry results and many artifacts. (b) CR-GAN generates realistic images of different styles.}
	\label{fig:DR-GAN-ours-random}
	\vspace{-2mm}
\end{figure}

\textbf{Generative Ability}. We utilize DR-GAN and CR-GAN to generate images from random noises. In Fig.~\ref{fig:DR-GAN-ours-random}, CR-GAN can produce images with different styles, while DR-GAN leads to blurry results. This is because the single-pathway generator of DR-GAN learns incomplete representations in the embedding space, which fails to handle random inputs. Instead, CR-GAN produces favorable results with complete embeddings. 

\subsection{Comparison with BiGAN}
\begin{figure}[t]
	\centering
	\includegraphics[width=\linewidth]{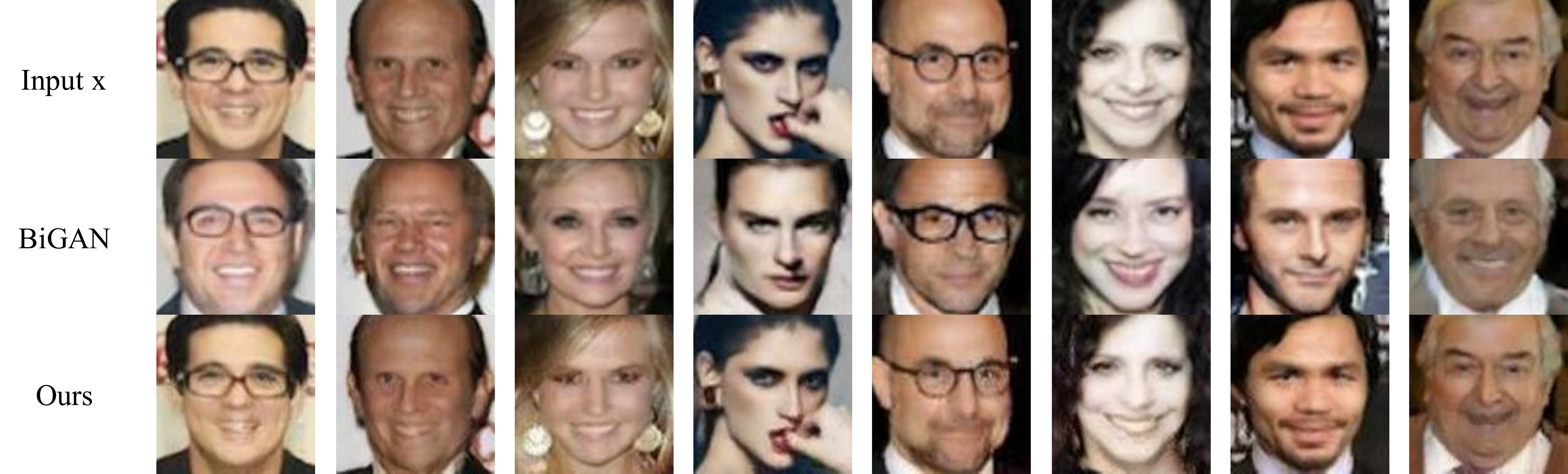}
	\caption{Reconstruction results on CelebA. BiGAN (Row 2) cannot keep identity well. Ours (Row 3) produces better results.}
	\label{fig:bigan}
	\vspace{-4mm}
\end{figure}

To compare our method with BiGAN, we qualitatively show the image reconstruction results of both methods on CelebA in Fig.~\ref{fig:bigan}. We can find that as demonstrated by~\cite{donahue2016adversarial,dumoulin2016adversarially}, BiGAN cannot reconstruct the data correctly, while CR-GAN keeps identities well due to the explicit reconstruction loss we employed.

\section{Conclusion}
In this paper, we investigate learning ``complete representations" of GAN models. We propose CR-GAN that uses a two-pathway framework to achieve the goal. Our method can leverage both labeled and unlabeled data for self-supervised learning, yielding high-quality multi-view image generations from even ``unseen'' data in wild conditions.

\section*{Acknowledgements}
This work is partly supported by the Air Force Office of Scientific Research (AFOSR) under the Dynamic Data-Driven Application Systems program, NSF CISE program, and NSF grant CCF 1733843.


\end{document}